\documentclass{article}

\usepackage[preprint]{neurips_2022}

\usepackage[utf8]{inputenc} 
\usepackage[T1]{fontenc}    
\usepackage{hyperref}       
\usepackage{url}            
\usepackage{booktabs}       
\usepackage{amsfonts}       
\usepackage{nicefrac}       
\usepackage{microtype}      
\usepackage{xcolor}         

\usepackage{amsmath}  
\usepackage{graphicx}  
\usepackage{booktabs}  
\usepackage{dsfont}  
\usepackage{makecell}  
\usepackage{subcaption} 

\title{Towards mapping the contemporary art world with ArtLM: an art-specific NLP model}

\author{
Qinkai Chen$^{1}$,
Mohamed El-Mennaoui$^{2}$,
Antoine Fosset$^{2}$,
Amine Rebei$^{2}$\thanks{Corresponding author \texttt{amine.rebei@docent-art.com}}\\
\textbf{
Haoyang Cao$^{1}$,
Philine Bouscasse$^{2}$,
Christy Eóin O'Beirne$^{3}$,
Sasha Shevchenko$^{3}$,
Mathieu Rosenbaum$^{1,2}$}\\
$^{1}$CMAP, Ecole Polytechnique, France $^{2}$Docent Art, France
$^{3}$Goldsmiths CCA, UK
}

\begin{document}

\maketitle

\begin{abstract}
  With an increasing amount of data in the art world, discovering artists and artworks suitable to collectors' tastes becomes a challenge. It is no longer enough to use visual information, as contextual information about the artist has become just as important in contemporary art.
  In this work, we present a generic Natural Language Processing framework (called ArtLM) to discover the connections among contemporary artists based on their biographies.
  In this approach, we first continue to pre-train the existing general English language models with a large amount of unlabeled art-related data. We then fine-tune this new pre-trained model with our biography pair dataset manually annotated by a team of professionals in the art industry.
  With extensive experiments, we demonstrate that our ArtLM achieves 85.6\% accuracy and 84.0\% F1 score and outperforms other baseline models.
  We also provide a visualization and a qualitative analysis of the artist network built from ArtLM's outputs.
  
\end{abstract}

\section{Introduction}

The contemporary art market has been growing at an unprecedented pace since the beginning of this century \citep{kraussl2016there}.
Traditionally, collectors rely on social interactions, professional advice and media to find the artworks they wish to acquire.
However, with more artists and their artworks on the market, this process of recommending suitable artworks to collectors with different tastes became inefficient.
Recently, with the wave of accelerated digitization in the art industry due to the COVID-19 pandemic \citep{noehrer2021impact} and the rapid development in artificial intelligence, researchers started to modernize this art discovery process with state-of-the-art technologies.

Most researchers focus on the visual aspect of the artwork. \citet{elgammal2018shape} fine-tune some commonly used model structures in computer vision, such as AlexNet \citep{krizhevsky2017imagenet} and ResNet \citep{he2016deep}, to classify the styles of the paintings.
\citet{kim2018finding} further expand this method to group artworks into pre-defined art concepts and principles in multiple dimensions. However, the visual information has certain limits as contemporary art goes beyond the aesthetics. The surrounding narration is also essential: the artist practice, the message conveyed in the work, the influences used by the artist and many more contextual attributes also have a strong influence on the taste in art.

Hence, the easily available textual information, such as the artists' biographies, comes to our attention.
Few existing research in the literature focus on this domain. \citet{kim2022formal} transform style labels into a fixed-length embedding with the help of Natural Language Processing (NLP) models and combine it with visual elements.
\citet{fosset2022docent} tag the artworks by analyzing the authors' biographies through static (Word2Vec \citep{mikolov2013efficient}) or contextualized (BERT \citep{devlin2018bert,chen2021stock}) embeddings without supervised fine-tuning or unsupervised pre-training.

To close the gap in the existing research introduced in Section \ref{sec:related_work}, adopting the idea of transfer learning \citep{weiss2016survey}, we introduce a generic NLP framework that we call ArtLM to solve this artist discovery problem formulated in Section \ref{sec:problem_formulation}.
The details of ArtLM are introduced in Section \ref{sec:identifying_artist_connection_with_artlm}.
In ArtLM, we first reuse a general English language model pre-trained on a very large amount of English texts.
We then continue the pre-training process with a large non-labeled art-related text dataset from WikiArt\footnote{https://www.wikiart.org/} in order to add art knowledge to the general English model.
We finally fine-tune this pre-trained art model with a small set of biography pairs manually labeled by a team of professionals in the contemporary art industry (an example is shown in Table \ref{tab:pair_example}). This last step makes our model specialized in this particular artist pair classification task.

\begin{table}[htbp]
  \centering
  \tiny
  \caption{An example of our labeled artist biography pair data.
  \textit{bio\_a} is the biography of the \textit{artist\_a} and \textit{bio\_b} is the biography of the \textit{artist\_b}. The \textit{label} is manually annotated by a team of professionals in the art industry signifying if two artists are connected (1 if connected and 0 if not connected).
  Our goal is to train a model which can tell if two artists are connected given their biographies.}
    \begin{tabular}{p{3em}p{3em}p{23em}p{23em}c}
    \toprule
    \multicolumn{1}{c}{artist\_a} & \multicolumn{1}{c}{artist\_b} & \multicolumn{1}{c}{bio\_a}  & \multicolumn{1}{c}{bio\_b} & \multicolumn{1}{c}{label} \\
    \midrule
    Carl-Edouard Keita & Kudzanai-Violet Hwami & Born in 1992 in Abidjan, Carl-Edouard Keita now lives and works in New York. A 2021 graduate of the New York Academy of Art, Carl-Edouard Keita also won the prize for best draughtsman for his graduation work, some of which is presented in this group exhibition. Carl-Edouard Keita discovered the history of African art during his economics studies in Atlanta, through a course offered at his university. As he describes it himself, this discovery was a real aesthetic revelation for him. & Having fled her homeland due to the political unrest and turmoil when she was a child, Zimbabwe-born painter Kudzanai-Violet Hwami expresses her personal experiences of dislocation, displacement and fragmentation through her striking figurative paintings. The artist is interested in the collapsing of geography and time and space symptomatic of a globalised world and high-speed internet, through which both people and information can travel quickly. & 1 \\
    \midrule
    Sun Xun & Cheng Xinhao & Sun Xun was born in 1980 in Fuxin, Liaoning province, China. Currently lives and works in Beijing. Recent and past histories, intransigent conflicts and tensions, sequential flashes of hand-created images of these are the irrevocable features of Sun Xun's artistic practice that fuses the line between art and animation. A graduate from the Printmaking Department of the China Academy of Arts in 2005, Sun Xun was a professor at the prestigious Academy before founding in 2006 his own Animation Studio. His work primarily involves making images using various materials such as colour powder, woodcuts and traditional ink, and collating these to produce a film, which is often presented in an immersive setting. & Cheng Xinhao (b.1985, Yunnan, China). After receiving his PhD on Chemistry from Peking University in 2013, Cheng continued his career as a photographer, investigating on the issues in the modernization, the construction of knowledge as well as the production of space in Chinese society.  & 0 \\
    \bottomrule
    \end{tabular}%
  \label{tab:pair_example}%
\end{table}%

In section \ref{sec:experiments}, with extensive experiments, we demonstrate that our ArtLM approach outperforms other baseline models in both accuracy and F1 score. We also prove that the continued pre-training and the fine-tuning are both essential to the state-of-the-art performance, and that these two steps are robust to the base model choices.
In addition, we visualize the artist network built from our model output and compare it with the ground-truth.

\section{Related Work}
\label{sec:related_work}

\subsection{Artwork Discovery and Recommendation}
Recently, art discovery and application of recommendation systems to the contemporary art scene attracted attention from the research community. For example, \citet{messina2018expl} study the impact of including metadata of artworks in addition to visual features on artwork recommendation.
With images and transaction data from \textit{UGallery}\footnote{https://www.ugallery.com/} online artwork store, they show that a hybrid approach improves the performance of the recommendation.

\citet{fosset2022docent} propose a novel and innovative approach to build a recommendation engine suited for contemporary art complexities. They combine visual attributes of artworks and contextual information about artists to build similarity graphs, allowing them to make a global and informed recommendation close to users' artistic tastes. However, for the contextual aspect in this work, the authors rely mainly on static Word2Vec embeddings to tag text data with different aspects conveyed by the artists, such as themes, subjects, emotions, etc.

More recently, \citet{wang2022ssar} use Named Entity Recognition (NER) to extract the topics from the texts from Wikipedia and Graph Convolutional Networks (GCN) to build a artists graph. They further build a recommendation framework (SSAR-GNN) based on the relationship among the artists.

\subsection{Language Model}
\label{subsec:language_model}

In Natural Language Processing, a Language Model (LM) refers to a model that can represent the probability distribution over sequences of words in a language.
LMs are generally trained on very large corpus with unsupervised learning tasks.
They are usually used as general-propose models and can be adapted to various downstream tasks, including sentence pair classification, through transfer learning \citep{weiss2016survey}.

There are two main types of LM in modern NLP: Masked Language Modeling (MLM) \citep{taylor1953cloze,devlin2018bert} and Causal Language Modeling (CLM) \citep{radford2018improving}.
The main difference between these two types of models is the training target.
MLM aims at predicting a randomly masked word in a sentence given the context before and after the word.
This target is used in LMs such as BERT, RoBERTa \citep{liu2019roberta}, etc.
In contrary, CLM's target is to predict the next word given its preceding words, such as GPT \citep{radford2019language}. There are also models which use both training targets, for instance, XLNet \citep{yang2019xlnet} and XLM \citep{lample2019cross}.

Another commonly used training target is Next Sentence Prediction (NSP), which is a binary classification loss for predicting whether two segments follow each other in the original text. The NSP task is usually combined with other tasks such as MLM or CLM, although some researchers argue that the NSP task does not have a clear contribution in the language modeling process \citep{joshi2019bert,lample2019cross}.

In this study, we use MLM as our training target since MLM shows better results when the goal is to learn a good representation of the input documents while CLM is more advantageous for text generation tasks such as machine translation and chatbots \citep{eo2021comparative}. We also choose different base models with and without the NSP loss to demonstrate the usefulness of this task.

\subsection{Natural Language Processing in Art}

NLP has already been proven to be useful and is widely used in many domains, including machine translation \citep{stahlberg2020neural}, finance \citep{chen2022graph}, music \citep{oramas2018natural}, etc.
However, in the visual art industry, previous research mostly focuses on the visual aspect of the artworks and ignore the rich information in the texts associated with the artists and the artworks.

There are several early attempts in applying NLP technologies on artworks, such as \citet{kim2022formal} and \citet{fosset2022docent}.
Nevertheless, both works only use the language models to generate embeddings from the texts and combine them with the visual elements, without training a task-specific model.
To the best of our knowledge, there is no previous work which focuses on discovering artist relationships based on their biographies.

\section{Problem Formulation}
\label{sec:problem_formulation}

We formulate this artist connection discovery problem as a sentence pair binary classification problem.
Our goal is to determine if two artists are connected given their respective biographies.

Suppose that we have two artists $A_{i}$, $A_{j}$ and their biographies are denoted by $B_{i}$, $B_{j}$.
A team of art professionals can determine if they are connected based on different aspects, including background, themes, style, techniques, etc.
We use $Y_{i, j}$ to denote this ground-truth, with
\begin{equation*}
    Y_{i, j} = 
    \begin{cases}
      1 & \text{if $A_{i}$ and $A_{j}$ are connected} \\
      0 & \text{otherwise} \\
    \end{cases}.
\end{equation*}

However, there are thousands of artists in our artist database and it is impossible to determine all relationships through manual annotation.
Hence, in this study, we are interested in building the connections among them automatically based on their biographies, which comprise the information of the artists in different dimensions. 
We can describe this process as
\begin{equation*}
    p_{i, j} = f(B_{i}, B_{j}, \theta)
\end{equation*}
where $p_{i, j}$ denotes our predicted probability of a connection between $A_{i}$ and $A_{j}$.
$f$ is our prediction model and $\theta$ denotes the trainable parameters.

Given $f$, our goal is to find $\theta$ which minimizes the cross-entropy loss \citep{good1992rational} defined as
\begin{equation*}
    \label{eq:loss}
    -\sum_{i, j} Y_{i, j} \cdot \log p_{i, j} + (1 - Y_{i, j}) \cdot \log (1-p_{i, j}).
\end{equation*}

\section{Identifying Artist Connection with ArtLM}
\label{sec:identifying_artist_connection_with_artlm}

There are three main components in our artist connection discovery method:
\begin{enumerate}
    \item (unsupervised) pre-train a LM with generic English corpus
    \item (unsupervised) continue to pre-train the LM with the same goal, but with art-related texts to get a LM with domain-specific knowledge in art (ArtLM)
    \item (supervised) fine-tune the ArtLM with labeled artist biography pairs
\end{enumerate}

\begin{figure}[h]
    \centering
    \includegraphics[width=\textwidth]{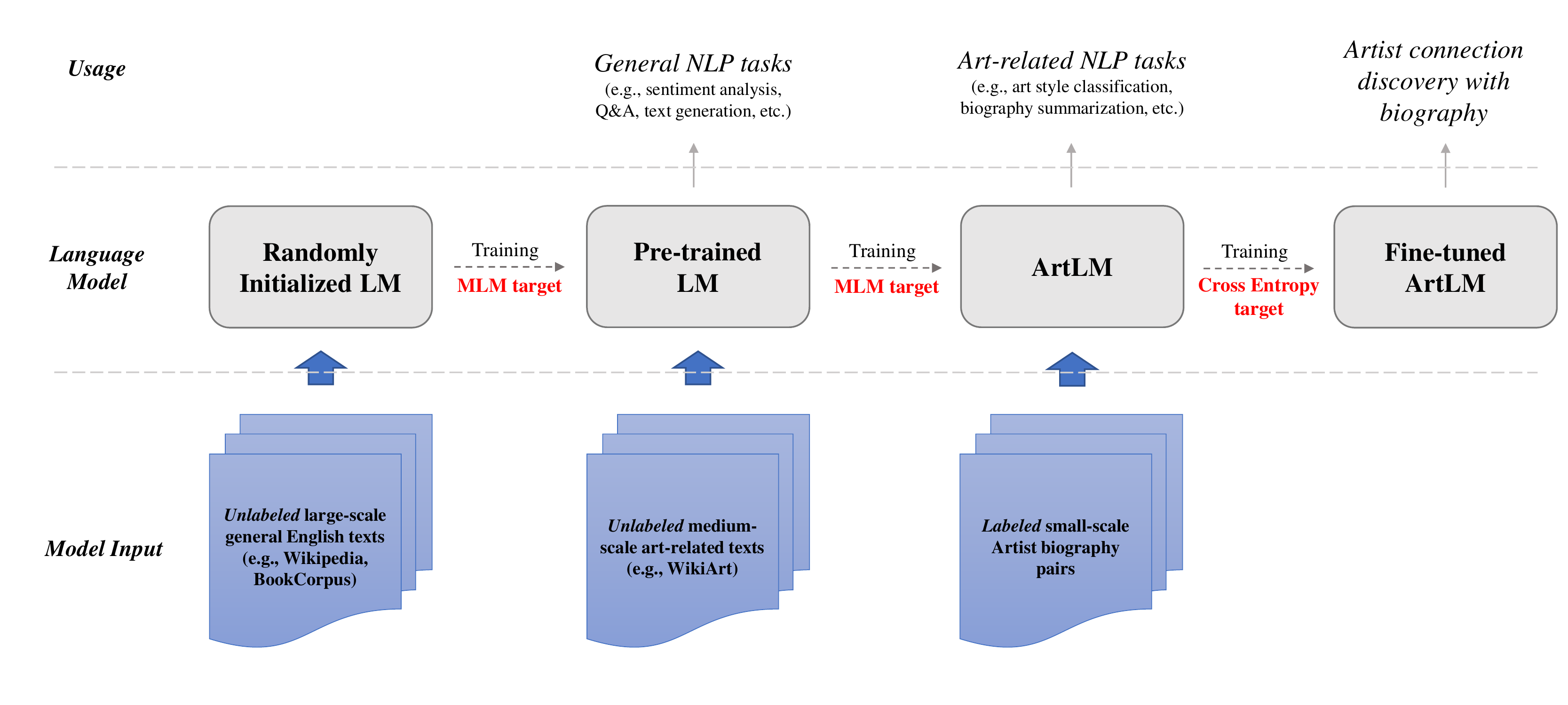}
    \caption{Overview of our artist connection discovery method, including two unsupervised pre-training phases and one supervised fine-tuning phase.}
    \label{fig:procedure}
\end{figure}

The method is illustrated in Figure \ref{fig:procedure} and we introduce each component in detail in the following subsections.

\subsection{Pre-trained LM}

At first, we want to learn the general characteristics of a language as our base model, which we refer as the pre-trained base language model. 

In order to generalize, this model is usually trained on a very large corpus. For example, the BERT model is trained on BookCorpus \citep{mcenery2006corpus} and English Wikipedia which have more than 3.3 billion words in total. Hence, this process is computationally expensive and time consuming. In our experiments, we use publicly available pre-trained English LMs instead of training from scratch by ourselves.

To prove that our approach is generic and not model-dependant, we adopt three different pre-trained LMs of different sizes and training targets as our base models, including DistilBERT, BERT and RoBERTa. Their details are shown in Table \ref{tab:pre_trained_models}.

\begin{table}[h]
  \centering
  \caption{Details of the pre-trained models used in our experiments}
    \begin{tabular}{ccccc}
    \toprule
    model & size  & \makecell{\# params. \\ (millions)} & training target & \makecell{training dataset \\ size} \\
    \midrule
    \makecell{DistilBERT-based-uncased \\ \citep{sanh2019distilbert}} & small & 66 & MLM, NSP & 16GB \\
    \makecell{BERT-base-uncased \\ \citep{devlin2018bert}} & large & 110 & MLM, NSP & 16GB \\
    \makecell{RoBERTa-base \\ \citep{liu2019roberta}} & large & 125 & MLM & 160GB \\
    \bottomrule
    \end{tabular}%
  \label{tab:pre_trained_models}%
\end{table}%

\subsection{ArtLM}
\label{subsec:artlm}

\citet{howard2018universal} show that further pre-training a LM with domain-specific data helps improve the performance. For example, \citet{araci2019finbert} and \citet{chen2021stock} further pre-train BERT model with financial corpus, \citet{wada2020pre} applies this methodology on medical texts, etc.

Following the same idea, we reuse the weights in the pre-trained base models and continue to train these models with MLM target (Section \ref{subsec:language_model}) with a large amount of art-related texts.
Concretely, it means that we randomly replace around 15\% of the words in the sentences with a special character \texttt{[MASK]} and we ask the model to predict this word. We minimize the sum of cross-entropy loss of all the masked characters using stochastic gradient descendant (SGD) \citep{robbins1951stochastic}.
We use ArtLM (Art Language Model) to denote this derived model.

In our case, we use the texts from WikiArt for this propose.
We introduce this dataset in detail in Section \ref{subsubsec:wikiart_data}.

\subsection{ArtLM Fine-tuning}
\label{subsec:artlm_fine_tuning}

Starting from a language model, we can continue to fine-tune it with labeled data to get a model specializing in our task: artist biography pair classification.
As recommended by \citet{devlin2018bert}, we construct our labeled dataset in the format of $\texttt{[CLS]} \textrm{Trunc}(B_{i}) \texttt{[SEP]} \textrm{Trunc}(B_{j})$, where \texttt{[CLS]} denotes the special character for class labels (0 or 1 in our case) and \texttt{[SEP]} represents the special character separating two biographies.
\textit{Trunc} is the truncating function which removes the tailing words if the number of words in a biography exceeds the maximum limit.
This is to ensure that two biographies are homogeneous with similar characteristics.

In this step, we add another fully-connected layer and a softmax function\footnote{$\textrm{softmax}(\vec{z})_{i} = \frac{e^{z_{i}}}{\sum_{j=1}^{K} e^{z_{j}} }$} at the end of the model.
This layer takes the vector representing \texttt{[CLS]} as input and outputs $p_{i, j}$, the predicted probability of a connection between artists $A_{i}$ and $A_{j}$.
We train this fully-connected layer and all other layers in the model jointly with the target function shown in Equation \ref{eq:loss}.
Our final prediction can therefore be written as
\begin{equation*}
    \widehat{Y_{i, j}} = \mathds{1}_{p_{i, j} > 0.5}
\end{equation*}
where $\mathds{1}$ denotes the indicator function.

\section{Experiments}
\label{sec:experiments}

\subsection{Datasets}

In this study, we use two different datasets to pre-train and fine-tune our ArtLM.
We use a large dataset scraped from WikiArt to add art-related knowledge to the pre-trained base model and we use a smaller manually annotated artist biography pair dataset to fine-tune this model.
We introduce the details of these two datasets in the following subsections.

\subsubsection{WikiArt Data}
\label{subsubsec:wikiart_data}
WikiArt is an online art encyclopedia that contains art-related text data. We only use artists' biographies in this database, which is the most relevant to our task.


There are 3,110 articles scrapped from the WikiArt database, accounting for more than 1.3 million tokens in total.

\subsubsection{Labeled Artist Biography Pair Data}

To build the annotated biography pair data, we select about 850 contemporary artists spanning different artistic movements, techniques and expressing a variety of subjects and themes through their artworks. This leads to around 360,000 possible pairs in total, which is too large to annotate manually. To sample a smaller dataset for manual labeling, we first use the baseline model to get a set of potential connections. We then randomly sample a subset with around 1,500 pairs.
We ask art curators from a renowned university to label these pairs as 0 if no relationship between the artists can be drawn based on the biographies, and label as 1 otherwise.

An example of the labeled data is already shown in Table \ref{tab:pair_example}.
We provide some statistics of the dataset in Table \ref{tab:stats_biography}.

The average number of words of all the biographies is 191, while the maximum number of words can reach 2,000. If there are more than 255 words in a sentence, we truncate the tailing words as mentioned in Section \ref{subsec:artlm_fine_tuning} to ensure that the sample fits in the model.

\begin{table}[htbp]
  \centering
  \caption{Some statistics of our labled artist biography pair data}
    \begin{tabular}{ccccc}
    \toprule
    label & meaning & count & proportion & word count \\
    \midrule
    0     & not connected & 767   & 52.6\% & 315k \\
    1     & connected & 691   & 47.4\% & 242k \\
    \midrule
    \multicolumn{2}{c}{total} & 1,458   & 100.0\% & 557k \\
    \bottomrule
    \end{tabular}%
  \label{tab:stats_biography}%
\end{table}%

\subsection{Baseline Models}
\label{subsec:baseline_models}

In order to prove the effectiveness of our ArtLM approach, we compare its performance with other baseline models. We include the following baseline models.

\paragraph{Random Guess}
We randomly assign the label 0 or 1 to each biography pair, with probability of $P_{0}$ and $P_{1}$ respectively.
We test two scenarios: $P_{0} = 1$ (all pairs are predicted as 0) and $P_{0} = 0.5$ (the proportion of the label 0 in the ground-truth).
We use the expression \texttt{Random Guess - $P_{0}$} to denote the random guess prediction results with different probabilities.

\paragraph{Static embedding from Word2Vec}
We use Word2Vec embeddings to create associations between artists biographies and themes based on keywords from a pre-defined list of themes. We consider that two artists are linked if they have at least one tag in common.

\paragraph{Contextualized embedding from BERT}
Instead of using static embeddings from Word2Vec, we use contextualized embeddings generated from the last layer of BERT.

\paragraph{Fine-tuned base models}

In addition to the above baseline models, we also report the results from fine-tuning base models without continuous pre-training on WikiArt data (we skip the step 2 mentioned in Section \ref{sec:identifying_artist_connection_with_artlm}). This intermediate result helps demonstrate the performance gain from pre-training a ArtLM instead of directly using the pre-trained base models.

In the following sections, we use the prefix \texttt{FT-Base} to denote the results from a fine-tuned base model and the prefix \texttt{FT-Art} to denote the results from a fine-tuned ArtLM. For example, two BERT variants are respectively denoted by \texttt{FT-Base-BERT} and \texttt{FT-Art-BERT}.
We report the performance from all three base models mentioned in Table \ref{tab:pre_trained_models}.

\subsection{Experiment Setup}

\subsubsection{Experiment Method}

For the ArtLM pre-training process (Section \ref{subsec:artlm}), we randomly mask 15\% of the words of all the sentences included in the WikiArt articles and continue to pre-train from the last checkpoint of one of the pre-trained base models. For all base models, we set the learning rate to 2e-5 and iterate on the dataset for 3 epochs.

In contrary, for the ArtLM fine-tuning process (Section \ref{subsec:artlm_fine_tuning}),
we split our data into 5 folds randomly for a cross-validation, since the size of our manually annotated data is relatively small. It means that we split the data equally into 5 parts, and we fine-tune 5 different models with one part being the test set and the remaining 4 parts being the training set and validation set.
This helps avoid the bias caused by data split. In this fine-tuning process, we set the learning rate to 2e-5 and iterate on the dataset for 20 epochs. We select the best epoch according to the F1 score on the validation set and perform inference on the test set.

We use the mean and the standard deviation of the accuracy and F1 score of 5 folds to evaluate the model.
Given a confusion matrix $\begin{pmatrix}
  tp & fn\\ 
  fp & tn
\end{pmatrix}$\footnote{$tp$, $fn$, $fp$ and $tn$ denote true positive, false negative, false positive and true negative respectively.},
the accuracy is defined as
\begin{equation*}
    Acc = \frac{tp + tn}{ tp + tn + fp + fn}
\end{equation*}
and the F1 score is defined as
\begin{equation*}
    F_{1} = \frac{2tp}{2tp + fp + fn}.
\end{equation*}

\subsubsection{Hardware}

The experiments are conducted on a Nvidia Tesla V100 GPU with 16 GB memory and 900 GB/s bandwidth.
The machine also has 8 cores of Intel Xeon E5-2686 CPU for other computationally inexpensive jobs. An ArtLM pre-training process usually takes 10-20 minutes to finish depending on the base model size, while an ArtLM fine-tuning process generally takes 30-60 minutes for all 5 folds.

\subsection{Experiment Results}
\label{subsec:experiment_results}

The detailed results of different ArtLM variants and the baseline models mentioned in Section \ref{subsec:baseline_models} are shown in Table \ref{tab:results}.

\begin{table}[h]
  \centering
  \caption{Experiment results of the ArtLM variants and other baseline models. If applicable, the numbers of accuracy and F1 score in this table are shown in the format: average of 5 folds $\pm$ standard deviation of 5 folds. The fine-tuning time is calculated as the mean on 5 folds. We note that no training is involved in the embedding baseline methods, there is therefore no cross-validation and the accuracy and F1 score are calculated based on the whole dataset.}
    \begin{tabular}{ccccc}
    \toprule
    model & accuracy & F1    & \makecell{fine-tuning time \\ (min. / fold)} & \makecell{pre-training \\ time (min.)} \\
    \midrule
    Random Guess - 1 & 0.500 & 0.000 & - & - \\
    Random Guess - 0.5 & 0.500 & 0.500 & - & - \\
    \midrule
    Word2Vec embedding & 0.539  & 0.196  & -    & - \\
    BERT embedding &  0.581 & 0.320 & -    & - \\ 
    \midrule
    FT-Base-DistilBERT & 0.838 $\pm$ 0.016 & 0.826 $\pm$ 0.019 & 6     & - \\
    FT-Base-BERT & 0.844 $\pm$ 0.022 & 0.832 $\pm$ 0.020 & 12    & - \\
    FT-Base-RoBERTa & 0.844 $\pm$ 0.021 & 0.824 $\pm$ 0.021 & 12    & - \\
    \midrule
    FT-Art-DistilBERT & 0.846 $\pm$ 0.020 & 0.826 $\pm$ 0.019 & 6     & 10 \\
    FT-Art-BERT & \textbf{0.856 $\pm$ 0.023} & \textbf{0.840 $\pm$ 0.021} & 12    & 18 \\
    FT-Art-RoBERTa & 0.854 $\pm$ 0.024 & 0.834 $\pm$ 0.826 & 12    & 21 \\
    \bottomrule
    \end{tabular}%
  \label{tab:results}%
\end{table}%

First, we note that all fine-tuned models outperform random guess and the baseline models based on embeddings.
This phenomenon demonstrates that the task-specific information added during the fine-tuning process helps the model better predict the connections between the artists.

Secondly, comparing the results from the fine-tuned base models and the fine-tuned ArtLMs, the ArtLM outperforms the base model on all three cases.
This proves the effectiveness of adding domain-specific information through continuing the pre-training process on a large amount of art-related texts. It also demonstrates that our ArtLM approach is effective on all base models and is robust to the model choices.

Finally, we observe that among the three ArtLM variants, \texttt{FT-Art-BERT} has the best performance, surpassing the both smaller \texttt{FT-Art-DistilBERT} and the larger \texttt{FT-Art-RoBERTa}, although RoBERTa reports a superior performance than BERT on most of the NLP tasks \citep{liu2019roberta}.
A possible cause is that the RoBERTa model removes the NSP task in the pre-training process, as mentioned in Section \ref{subsec:language_model} and in Table \ref{tab:pre_trained_models}.
However, our artist pair classification usage is similar to the idea behind the NSP task, which leads to the performance degradation after its removal.
Another possible explanation is that the RoBERTa is trained on a much larger corpus than BERT, it may require more art-related data in this second pre-training process in order to be significant.

\subsection{Artist Network}
\label{sec:artist_network}

\begin{figure}[h]
\centering
\begin{subfigure}[t]{0.45\linewidth}
  \centering
  \includegraphics[width=\linewidth]{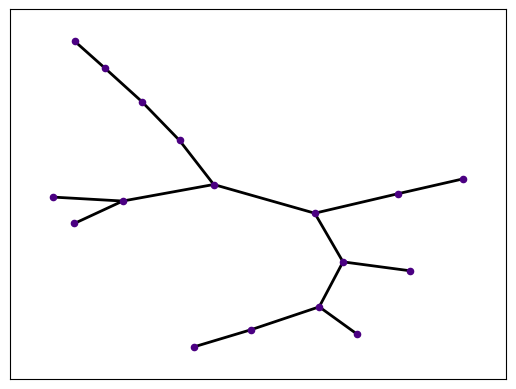}
  \caption{a sub-network of the ground-truth network}
  \label{subfig:label}
\end{subfigure}%
\begin{subfigure}[t]{0.45\linewidth}
  \centering
  \includegraphics[width=\linewidth]{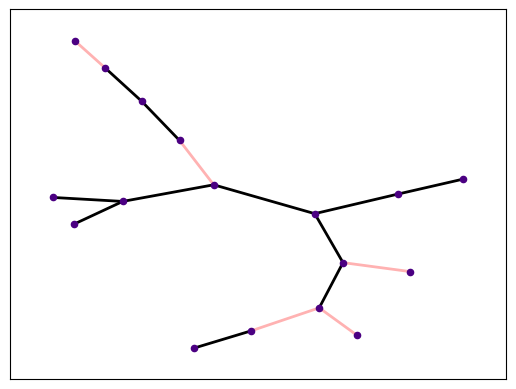}
  \caption{\texttt{FT-Art-BERT} prediction of (a)}
  \label{subfig:pred}
\end{subfigure}
\caption{Artist network. Figure (a) is a sub-network of the artist network built from the ground-truth. We perform analysis on this sub-network. Figure (b) is the prediction of (a) made from our \texttt{FT-Art-BERT} model.}
\label{fig:artist_network}
\end{figure}

With the prediction results in Section \ref{subsec:experiment_results}, we can build artist networks which can be used to visualize their relationships or to recommend related artists based on collector's interests in further studies.
We visualize some artist networks in Figure \ref{fig:artist_network}. In each sub-figure, a node denotes an artist. An edge between two nodes signifies that the two authors are connected, either from the ground-truth or from the prediction.
We first construct an artist network using the labeled ground-truth data.
We then zoom into one of the sub-graphs (Figure \ref{subfig:label}) for a qualitative analysis.

Figure \ref{subfig:pred} is the prediction of Figure \ref{subfig:label} made from our \texttt{FT-Art-BERT} model. In this figure, black edge denote the correctly predicted edges, red edges denote the edges that are present in the ground-truth but identified as not connected by \texttt{FT-Art-BERT}. Blue edges represent the relations that are present in the predictions but not in the ground-truth.
We can see that the ArtLM helps rebuild the framework of ground-truth network, although there are mismatches on some biography pairs\footnote{The Graph Edit Distance (GED) \citep{gao2010survey} between two graphs is 7.}, especially in the tails of the graph.

A potential usage of this artist network is artwork recommendation.
For example, we can recommend the frames of the connected artists if one collector likes an artist.

\section{Conclusion}

In this work, we classify if two contemporary artists are closely connected through their biographies.
We introduce a transfer learning based approach: Art Language Model (ArtLM), which comprises an unsupervised pre-training step and a supervised fine-tuning step.
Through extensive experiments, we demonstrate that ArtLM outperforms other benchmark models. We also prove that both pre-training and fine-tuning are essential to a better performance, and that this procedure is generic and robust to the choices of the base models.

For future study, we seek to integrate the ArtLM's outputs into a recommender system \citep{resnick1997recommender} to further enhance the artwork discovery process. We can also combine the visual and textual aspects of the artworks in a more complex system.

We also plan to enrich our manually annotated biography pair dataset with the appreciations from different aspects. It means that instead of classifying a pair of artists in a binary manner, we label their relationships from different dimensions, for example, style, background, influence, etc. This problem will therefore become a multi-class classification problem and allow us to better understand the relationship between two artists.




\bibliographystyle{plainnat}
\bibliography{bibliography}





\end{document}